\relax
\PassOptionsToPackage{dvipsnames}{xcolor}
\documentclass[11pt]{article}
\usepackage{mtsummit2019}
\usepackage{times}
\usepackage{url}
\usepackage{latexsym}
\usepackage[small,bf]{caption} 
\usepackage{titlesec}
\usepackage{multirow}
\usepackage{dashrule}
\usepackage{booktabs}

\usepackage[]{xcolor}
\usepackage{subfig}
\usepackage[textsize=large,disable]{todonotes}
\usepackage{booktabs}
\usepackage{adjustbox}
\usepackage{algorithm}
\usepackage{algpseudocode}
\usepackage[inline]{enumitem}
\usepackage{amsmath,amssymb}
\usepackage{pgfplots}
\usepackage{tabularx} 
\usepackage{url}

\newcommand{\note}[4][]{\todo[author=#2,color=#3,size=\scriptsize,fancyline,caption={},#1]{#4}} 
\newcommand{\pamela}[2][]{\note[#1]{pamela}{cyan}{#2}}

\newcommand{\adi}[2][]{\note[#1]{adi}{green}{#2}}

\newcommand{\Pamela}[2][]{\pamela[inline,#1]{#2}\noindent}
\newcommand{\Adi}[2][]{\adi[inline,#1]{#2}\noindent}

\renewcommand{\vec}[1]{{\boldsymbol{\mathbf{#1}}}}   

\usepackage{dirtytalk}

\title{Character-Aware Decoder for Translation\\ 
into Morphologically Rich Languages}

\author{Adithya Renduchintala$^*$
\and Pamela Shapiro$^*$
\and Kevin Duh \and Philipp Koehn\\
  Department of Computer Science\\
  Johns Hopkins University\\
  {\tt \{adi.r,pshapiro,phi\}@jhu.edu kevinduh@cs.jhu.edu}}
\date{}
\begin{document}
\maketitle
\begin{abstract}
Neural machine translation (NMT) systems operate primarily on words (or subwords), ignoring lower-level patterns of morphology.
We present a \emph{character-aware} decoder 
designed to capture such patterns when translating into morphologically rich languages.
We achieve character-awareness by augmenting both the softmax and embedding layers of an attention-based encoder-decoder model with convolutional neural networks that operate on the spelling of a word.
To investigate performance on a wide variety of morphological phenomena, we translate English into $14$ typologically diverse target languages using the TED multi-target dataset.
In this low-resource setting, the character-aware decoder provides consistent improvements 
with BLEU score gains of up to $+3.05$. 
In addition, we analyze the relationship between the gains obtained and properties of the target language and find evidence that our model does indeed exploit morphological patterns.
\let\thefootnote\relax\footnote{$^*$Equal Contribution}
\setcounter{footnote}{0}
\end{abstract}
\section{Introduction}
Traditional attention-based encoder-decoder neural machine translation (NMT) models learn \emph{word-level} embeddings, with a continuous representation for each unique word type~\cite{bahdanau2014neural}. However, this results in a 
long tail of rare words for which we do not learn good representations.
More recently, it has become standard practice to mitigate the vocabulary size problem with Byte-Pair Encoding (BPE) \cite{Gage1994,sennrich2016subword}. 
BPE iteratively merges consecutive characters into larger chunks based on their frequency, which results in the breaking up of less common words into ``subword units.'' 

While BPE addresses the vocabulary size problem, the spellings of the subword units are still 
ignored.
On the other hand, purely \emph{character-level} NMT translates one character at a time and 
can implicitly learn about 
morphological patterns within words as well as generalize to unseen vocabulary.
Recently,~\newcite{cherry2018revisiting} show that very deep character-level models can outperform BPE, however, the smallest data size evaluated was 2 million sentences, so it is unclear if the results hold for low-resource settings and when translating into a range of different morphologically rich languages.
Furthermore, tuning deep character-level models is expensive, even for low-resource settings. Deep character-level models are sensitive to the dropout rate and tuning takes much longer due to longer sequence lengths~\cite{cherry2018revisiting}.
A middle-ground alternative is character-\emph{aware} word-level modeling.
Here, the NMT system operates over words but uses word embeddings that are sensitive to spellings 
and thereby has the ability to learn morphological patterns in the language.
Such character-aware approaches have been applied successfully in NMT to the \emph{source-side} word embedding layer \cite{costa2016character}, 
but surprisingly, similar gains have not been achieved on the target side~\cite{belinkov2017neural}. 

While source-side character-aware models only need to make the \emph{source embedding layer} character-aware, on the target-side we require both the \emph{target embedding layer} and the \emph{softmax layer} to be character-aware, which presents additional challenges. 
We find that the trivial application of methods from~\newcite{costa2016character} to these target-side embeddings results in significant drop in performance. Instead, we propose mixing compositional and standard word embeddings via a gating function. While simple, we find it is critical to successful target-side character awareness.


It is worth noting that unlike some purely character-level methods our aim is not to generate novel words, though this method can function on top of subword methods which do so~\cite{shapiro2018bpe}. Rather, the character-aware representations decrease the sparsity of embeddings for rare words or subwords, which are a problem in low-resource morphologically rich settings.
We summarize our contribution as follows:
\begin{enumerate}
 \item We propose a method for utilizing character-aware embeddings in an NMT decoder that can be used over word or subword sequences.
\item We explore how our method interacts with BPE over a range of merge operations (including word-level and purely character-level) and highlight that there is no ``typical BPE'' setting for low-resource NMT.
\item We evaluate our model on $14$ target languages and observe consistent improvements over baselines. Furthermore, we analyze to what extent the success of our method corresponds to improved handling of target language morphology. 
\end{enumerate}
\section{Related Work}

NMT has benefited from character-aware word representations on the source side~\cite{costa2016character}, which follows language modeling work by \newcite{kim2016character} and generate source-side input embeddings using a CNN over the character sequence of each word.
Further analysis revealed that hidden states of such character-aware models have increased knowledge of morphology
~\cite{belinkov2017neural}. 
They additionally try using character-aware representations in the target side embedding layer, leaving the softmax matrix with standard word representations, and found no improvements. 

Our work is also aligned with the character-aware models proposed in \cite{kim2016character}, but we additionally employ a gating mechanism between character-aware representations and standard word representations similar to language modeling work by \cite{miyamoto2016gated}.
However, our gating is a learned type-specific vector rather than a fixed hyperparameter.

There is additionally a line of work on purely character-level NMT, which generates words one character at a time \cite{ling2015character,chung2016character,passban2018improving}.
While initial results here were not strong, \newcite{cherry2018revisiting} revisit this with deeper architectures and sweeping dropout parameters and find that they outperform BPE across settings of the merge hyperparameter. They examine different data sizes and observe improvements in the smaller data size settings---however, the smallest size is about 2 million sentence pairs. In contrast, we look at a smaller order of magnitude data size and present an alternate approach which doesn't require substantial tuning of parameters across different languages.

Finally, Byte-Pair Encoding (BPE) \cite{sennrich2016subword} has become a standard preprocessing step in NMT pipelines and provides an easy way to generate sequences with a mixture of full words and word fragments. 
Note that BPE splits are agnostic to any morphological pattern present in the language, for example the token \texttt{politely} in our dataset is split into \texttt{pol+itely}, instead of the linguistically plausible split \texttt{polite+ly}.\footnote{We observe this split when merge parameter was $15$k.}
Our approach can be applied to word-level sequences and sequences at any BPE merge hyperparameter greater than $0$.
Increasing the hyperparameter results in more words and longer subwords that can exhibit morphological patterns.
Our goal is to exploit these morphological patterns and enrich the word (or subword) representations with character-awareness.
\section{Encoder-Decoder NMT} \label{method}
An attention-based encoder-decoder network 
\cite{bahdanau2014neural,luong2015effective} models the probability of a target sentence $\vec{y}$ of length $J$ given a source sentence $\vec{x}$ as:
\begin{align}
p(\vec{y} \mid \vec{x}) &= \prod_{j=1}^{J}p(y_j \mid \vec{y}_{0:j-1}, \vec{x}; \vec{\theta}) \label{eq:objective}
\end{align}
where $\vec{\theta}$ represents all the parameters of the network. 
At each time-step the $j'$th output token is generated by:
\begin{align}
p(y_j \mid \vec{y}_{0:j-1}, \vec{x})=\text{softmax}(\vec{W_o} \vec{s}_j) \label{eq:outputdist}
\end{align}
where $\vec{s}_j \in \mathbb{R}^{D\times1}$ is the decoder hidden state at time $j$ and $\vec{W_o} \in \mathbb{R}^{\lvert\mathcal{V}\rvert \times D}$ is the weight matrix of the softmax layer, which provides a continuous representation for target words.
$\vec{s}_j$ is computed using the following recurrence:
\begin{align}
\vec{s}_j &= \text{tanh}(\vec{W_c}\;[\vec{c}_j;\vec{\tilde{s}}_j])\\
\vec{\tilde{s}}_j &= f([\vec{s}_{j-1}; \vec{w_s}^{y_{j-1}}; \vec{\tilde{s}}_{j-1}])
\end{align}
where $f$ is an LSTM cell.\footnote{Note that our notation diverges from \newcite{luong2015effective} so that $\vec{s}_j$ refers to the state used to make the final predictions.} 
$\vec{W_s}  \in \mathbb{R}^{\lvert\mathcal{V}\rvert \times E}$ is the target-side embedding matrix, which provides continuous representations for the previous target word when used as input to the RNN. Here, $\vec{w_s}^{y_{j-1}} \in \mathbb{R}^{1 \times E}$ is a row vector from the embedding matrix $\vec{W_s}$ corresponding to the value of $y_{j-1}$. $\mathcal{V}$ is the target vocabulary set, $D$ is the is the RNN size and $E$ is embedding size. Often these matrices $\vec{W_o}$ and $\vec{W_s}$ are tied. 

The context vector $\vec{c}_j$ is obtained by taking a weighted average over the concatenation of a bidirectional RNN encoder's hidden states.
\begin{align}
\vec{c}_j &= \sum_{i=1}^{{I}}\alpha_i,\vec{h}_i \\
\alpha_i & = \frac{\exp{(\vec{s}_j^T \vec{W_a} \vec{h}_i)}}{\sum_l \exp{(\vec{s}_j^T \vec{W_a} \vec{h}_l})}
\end{align}
The attention matrix $\vec{W_a} \in \mathbb{R}^{D\times H}$ is learned jointly with the model, multiplying with the previous decoder state and bidirectional encoder state $\vec{h}_i \in \mathbb{R}^{H\times1}$, normalized over encoder hidden states via the softmax operation.

\section{Character-Aware Extension}
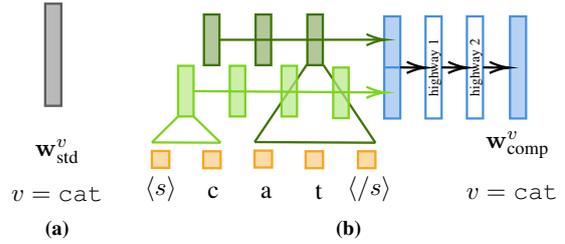
\begin{figure}
\centering
\resizebox{\columnwidth}{!}{  
\subfloat[]{
\tikzset{every picture/.style={line width=1.00pt}} 

\begin{tikzpicture}[x=0.75pt,y=0.75pt,yscale=-1,xscale=1]

\draw  [color={rgb, 255:red, 92; green, 92; blue, 92 }  ,draw opacity=1 ][fill={rgb, 255:red, 155; green, 155; blue, 155 }  ,fill opacity=0.7 ] (21,7.61) -- (30.33,7.61) -- (30.33,67.33) -- (21,67.33) -- cycle ;

\draw (28,94) node   {$\vec{w}^{v}_{\text{std}}$};
\draw (28,119) node   {$v=\texttt{cat}$};

\end{tikzpicture}

\label{fig:nocomposition}
}
\subfloat[]{

\tikzset{every picture/.style={line width=1.00pt}} 

\begin{tikzpicture}[x=0.75pt,y=0.75pt,yscale=-1,xscale=1]

\draw [color={rgb, 255:red, 65; green, 117; blue, 5 }  ,draw opacity=1 ]   (67.4,86.58) -- (137.52,86.98) ;

\draw [color={rgb, 255:red, 65; green, 117; blue, 5 }  ,draw opacity=1 ]   (97.8,41.2) -- (67.4,86.58) ;

\draw [color={rgb, 255:red, 65; green, 117; blue, 5 }  ,draw opacity=1 ]   (107.4,41.6) -- (137.52,86.98) ;

\draw  [color={rgb, 255:red, 245; green, 166; blue, 35 }  ,draw opacity=1 ][fill={rgb, 255:red, 245; green, 166; blue, 35 }  ,fill opacity=0.4 ] (7.8,91.78) -- (17.92,91.78) -- (17.92,101.67) -- (7.8,101.67) -- cycle ;
\draw  [color={rgb, 255:red, 245; green, 166; blue, 35 }  ,draw opacity=1 ][fill={rgb, 255:red, 245; green, 166; blue, 35 }  ,fill opacity=0.4 ] (37.6,91.98) -- (47.72,91.98) -- (47.72,101.87) -- (37.6,101.87) -- cycle ;
\draw  [color={rgb, 255:red, 245; green, 166; blue, 35 }  ,draw opacity=1 ][fill={rgb, 255:red, 245; green, 166; blue, 35 }  ,fill opacity=0.4 ] (67.4,91.58) -- (77.52,91.58) -- (77.52,101.47) -- (67.4,101.47) -- cycle ;
\draw  [color={rgb, 255:red, 245; green, 166; blue, 35 }  ,draw opacity=1 ][fill={rgb, 255:red, 245; green, 166; blue, 35 }  ,fill opacity=0.4 ] (97,91.98) -- (107.12,91.98) -- (107.12,101.87) -- (97,101.87) -- cycle ;
\draw  [color={rgb, 255:red, 245; green, 166; blue, 35 }  ,draw opacity=1 ][fill={rgb, 255:red, 245; green, 166; blue, 35 }  ,fill opacity=0.4 ] (127.4,91.98) -- (137.52,91.98) -- (137.52,101.87) -- (127.4,101.87) -- cycle ;
\draw  [color={rgb, 255:red, 74; green, 144; blue, 226 }  ,draw opacity=1 ][fill={rgb, 255:red, 74; green, 144; blue, 226 }  ,fill opacity=0.4 ] (142.6,13.08) -- (151.93,13.08) -- (151.93,72.8) -- (142.6,72.8) -- cycle ;
\draw  [color={rgb, 255:red, 126; green, 211; blue, 33 }  ,draw opacity=1 ][fill={rgb, 255:red, 184; green, 233; blue, 134 }  ,fill opacity=0.7 ] (23,42) -- (32,42) -- (32,71.6) -- (23,71.6) -- cycle ;
\draw  [color={rgb, 255:red, 126; green, 211; blue, 33 }  ,draw opacity=1 ][fill={rgb, 255:red, 184; green, 233; blue, 134 }  ,fill opacity=0.7 ] (53,42) -- (62,42) -- (62,71.6) -- (53,71.6) -- cycle ;
\draw  [color={rgb, 255:red, 126; green, 211; blue, 33 }  ,draw opacity=1 ][fill={rgb, 255:red, 184; green, 233; blue, 134 }  ,fill opacity=0.7 ] (83,42) -- (92,42) -- (92,71.6) -- (83,71.6) -- cycle ;
\draw  [color={rgb, 255:red, 126; green, 211; blue, 33 }  ,draw opacity=1 ][fill={rgb, 255:red, 184; green, 233; blue, 134 }  ,fill opacity=0.7 ] (113,43) -- (122,43) -- (122,72.6) -- (113,72.6) -- cycle ;
\draw  [color={rgb, 255:red, 65; green, 117; blue, 5 }  ,draw opacity=1 ][fill={rgb, 255:red, 65; green, 117; blue, 5 }  ,fill opacity=0.4 ] (38,12) -- (47,12) -- (47,41.6) -- (38,41.6) -- cycle ;
\draw  [color={rgb, 255:red, 65; green, 117; blue, 5 }  ,draw opacity=1 ][fill={rgb, 255:red, 65; green, 117; blue, 5 }  ,fill opacity=0.4 ] (68,12) -- (77,12) -- (77,41.6) -- (68,41.6) -- cycle ;
\draw  [color={rgb, 255:red, 65; green, 117; blue, 5 }  ,draw opacity=1 ][fill={rgb, 255:red, 65; green, 117; blue, 5 }  ,fill opacity=0.4 ] (98,12) -- (107,12) -- (107,41.6) -- (98,41.6) -- cycle ;
\draw [color={rgb, 255:red, 126; green, 211; blue, 33 }  ,draw opacity=1 ]   (32,56.8) -- (141.2,57.59) ;
\draw [shift={(143.2,57.6)}, rotate = 180.41] [color={rgb, 255:red, 126; green, 211; blue, 33 }  ,draw opacity=1 ][line width=1.00]    (10.93,-3.29) .. controls (6.95,-1.4) and (3.31,-0.3) .. (0,0) .. controls (3.31,0.3) and (6.95,1.4) .. (10.93,3.29)   ;

\draw [color={rgb, 255:red, 65; green, 117; blue, 5 }  ,draw opacity=1 ]   (47,26.8) -- (141.25,26.75) ;
\draw [shift={(143.25,26.75)}, rotate = 539.97] [color={rgb, 255:red, 65; green, 117; blue, 5 }  ,draw opacity=1 ][line width=1.00]    (10.93,-3.29) .. controls (6.95,-1.4) and (3.31,-0.3) .. (0,0) .. controls (3.31,0.3) and (6.95,1.4) .. (10.93,3.29)   ;

\draw [color={rgb, 255:red, 74; green, 144; blue, 226 }  ,draw opacity=1 ]   (142.6,42.94) -- (151.93,42.94) ;

\draw  [color={rgb, 255:red, 74; green, 144; blue, 226 }  ,draw opacity=1 ][fill={rgb, 255:red, 74; green, 144; blue, 226 }  ,fill opacity=0.4 ] (215.6,13.08) -- (224.93,13.08) -- (224.93,72.8) -- (215.6,72.8) -- cycle ;
\draw  [color={rgb, 255:red, 74; green, 144; blue, 226 }  ,draw opacity=1 ][fill={rgb, 255:red, 255; green, 255; blue, 255 }  ,fill opacity=1 ] (190.6,13.08) -- (199.93,13.08) -- (199.93,72.8) -- (190.6,72.8) -- cycle ;
\draw  [color={rgb, 255:red, 74; green, 144; blue, 226 }  ,draw opacity=1 ][fill={rgb, 255:red, 255; green, 255; blue, 255 }  ,fill opacity=1 ] (166.6,13.08) -- (175.93,13.08) -- (175.93,72.8) -- (166.6,72.8) -- cycle ;
\draw [color={rgb, 255:red, 126; green, 211; blue, 33 }  ,draw opacity=1 ]   (32,71.6) -- (47.72,86.98) ;

\draw [color={rgb, 255:red, 126; green, 211; blue, 33 }  ,draw opacity=1 ]   (23,71.6) -- (7.8,86.78) ;

\draw [color={rgb, 255:red, 126; green, 211; blue, 33 }  ,draw opacity=1 ]   (7.8,86.78) -- (47.72,86.98) ;

\draw [color={rgb, 255:red, 0; green, 0; blue, 0 }  ,draw opacity=1 ]   (151.93,42.94) -- (164.67,43.06) ;
\draw [shift={(166.67,43.08)}, rotate = 180.56] [color={rgb, 255:red, 0; green, 0; blue, 0 }  ,draw opacity=1 ][line width=1.00]    (10.93,-3.29) .. controls (6.95,-1.4) and (3.31,-0.3) .. (0,0) .. controls (3.31,0.3) and (6.95,1.4) .. (10.93,3.29)   ;

\draw [color={rgb, 255:red, 0; green, 0; blue, 0 }  ,draw opacity=1 ]   (175.93,42.94) -- (188.67,43.06) ;
\draw [shift={(190.67,43.08)}, rotate = 180.56] [color={rgb, 255:red, 0; green, 0; blue, 0 }  ,draw opacity=1 ][line width=1.00]    (10.93,-3.29) .. controls (6.95,-1.4) and (3.31,-0.3) .. (0,0) .. controls (3.31,0.3) and (6.95,1.4) .. (10.93,3.29)   ;

\draw [color={rgb, 255:red, 0; green, 0; blue, 0 }  ,draw opacity=1 ]   (199.93,42.94) -- (212.67,43.06) ;
\draw [shift={(214.67,43.08)}, rotate = 180.56] [color={rgb, 255:red, 0; green, 0; blue, 0 }  ,draw opacity=1 ][line width=1.00]    (10.93,-3.29) .. controls (6.95,-1.4) and (3.31,-0.3) .. (0,0) .. controls (3.31,0.3) and (6.95,1.4) .. (10.93,3.29)   ;

\draw (43.47,115.93) node  [align=left] {c};
\draw (104.67,115.53) node  [align=left] {t};
\draw (74.07,115.73) node  [align=left] {a};
\draw (172,43) node [rotate=-270] [align=left] {{\tiny highway 1}};
\draw (196,43) node [rotate=-270] [align=left] {{\tiny highway 2}};
\draw (134,114) node   {$\langle /s\rangle $};
\draw (13,114) node   {$\langle s\rangle $};
\draw (220,88) node   {$\vec{w}^{v}_{\text{comp}}$};
\draw (217,116) node   {$v=\texttt{cat}$};

\end{tikzpicture}

\label{fig:cnn}
}
}

\caption{Different approaches to generating embeddings. (a) standard word embedding that treats words as a single symbol. (b) CNN-based composition function. We use multiple CNNs with different kernel sizes over the character embeddings. The resulting hidden states are combined into a single word embedding via max pooling. Note that (b) shows only $2$ convolution filters 
for clarity, in practice we use 4.}
\label{fig:compositions}
\end{figure}

In this section we detail the incorporation of 
character-awareness into the two decoder embedding matrices $\vec{W_o}$ and $\vec{W_s}$. 
To begin, we consider an example target side word (or subword in the case of preprocessing with BPE), \texttt{cat}. 
In both $\vec{W_o}$ and $\vec{W_s}$, there exist row vectors, $\vec{w_o}^{\texttt{cat}}$ and $\vec{w_s}^{\texttt{cat}}$ that contain the continuous vector representation for the word \texttt{cat}.
In a traditional NMT system, these vectors are learned as the entire network tries to maximize the objective in Equation \ref{eq:objective}.
The objective does not require the vectors $\vec{w_o}^{\texttt{cat}}$ and $\vec{w_s}^{\texttt{cat}}$ to model any aspect of the spelling of the word. 
Figure~\ref{fig:nocomposition} illustrates a simple non-compositional word embedding. 

At a high level, we can view our notion of character-awareness as a composition function $\text{comp}(.;\vec{\omega})$, parameterized by $\vec{\omega}$, that takes the character sequence that makes up a word (i.e. its spelling) as input and then produces a continuous vector representation:
\begin{align}
\vec{w}_{\text{comp}}^{\texttt{cat}} &= \text{comp}(\langle s\rangle, \texttt{c},\texttt{a}, \texttt{t}, \langle/s\rangle; \vec{\omega}) \label{eq:comptemplate}
\end{align}
$\vec{\omega}$ is learned jointly with the overall objective. 
Special characters $\langle s \rangle$ and $\langle /s \rangle$ denote the beginning and end of sequence respectively.

Figure~\ref{fig:cnn} illustrates our compositional approach to generating embeddings~\cite{kim2016character}. First, a character-embedding layer converts the spelling of a word into a sequence of character embeddings. 
Next, we apply $4$ convolution operations, with kernel sizes $3,4,5$ and $6$, over the character sequence and the resulting output matrix is max-pooled.
We set the output channel size of each convolution to $\frac{1}{4}$ of the final desired embedding size.
The max-pooled vector from each convolution is concatenated to create the composed word representation. 
Finally, we add highway layers to obtain the final embeddings.

\subsection{Composed \& Standard Gating}\label{sec:CG}
The composition is applied to every type in the vocabulary and thus generates a complete embedding matrix (and softmax matrix). In doing so, we assume that \emph{every} word in the vocabulary has a vector representation that can be composed from its spelling sequence.
This is a strong assumption as many words, in particular high frequency words, are not normally compositional, e.g.\ the substring \texttt{ing} in \texttt{thing} is not compositional in the way that it is in \texttt{running}. 
Thus, we mix the compositional and standard embedding vectors. We expect standard embeddings to better represent the meaning of certain words, such has function words and other high-frequency words.
For each word $v$ in the vocabulary we also learn a gating vector $\vec{g}^{v} \in [0,1]^{1\times D}$.
\begin{align}
\vec{g}^{v} &= \sigma(\vec{w}_{\text{gate}}^{v}) \label{eq:gating}
\end{align}
Where, $\sigma$ is a sigmoid operation and type-specific parameters $\vec{w}_{\text{gate}}^{v}$ are jointly learned along with all the other parameters of the composition function. These parameters are regularized to remain close to $\vec{0}$ using dropout.
\footnote{However, in practice we found that this regularization did not affect performance noticeably in this setting.}
Our final mixed word representation for each word $v \in \mathcal{V}$ is given by:
\begin{align}
\vec{w}^{v}_{\text{mix}} = \vec{g}^{v} \odot \vec{w}^{v}_{\text{std}} + (\vec{1.} - \vec{g}^{v}) \odot \vec{w}^{v}_{\text{comp}} \label{eq:mixed}
\end{align}
Where $\vec{w}^{v}_{\text{mix}}$ is the final word embedding, $ \vec{w}^{v}_{\text{std}}$ is the standard word embedding, $\vec{w}^{v}_{\text{comp}}$ is the embedding by the composition function and $\vec{g}^{v}$ is the type-specific gating vector for the $v$'th word. The weight matrix is obtained by stacking the word vectors for each word $v \in \mathcal{V}$. 
The same representation is used for the target embedding layer and the softmax layer
i.e. we set $\vec{w_o}^{\texttt{cat}} = \vec{w_s}^{\texttt{cat}} = \vec{w}_{\text{mix}}^{\texttt{cat}}$, when $v = \texttt{cat}$. 
Thus, tying the composition function parameters for the softmax weight matrix and the target-side embedding matrix. 

Experiments comparing the standard embedding model and the compositional embedding model with and without gating are summarized in Table~\ref{tab:prelim}.
Row ``C'' shows the performance of naively using the  composition function (which works in the source-side) on the target-side.
We observe a catastrophic drop in BLEU ($-14.62$) compared to a standard NMT encoder-decoder.
The Character-aware gated model(CG), however, outperforms the baseline by $0.91$ BLEU points suggesting that the CNN composition function and standard embeddings work in a complementary fashion. 
\begin{table}[]
\small
\begin{center}
\begin{tabular}{@{}ll@{}}
\toprule
\textbf{Composition Method} & \textbf{BLEU} \\ \midrule
Std.\phantom{-} (no composition) & 26.84 \\
C\phantom{GS} (without gating)& 12.22 \\
CG (target embedding only) & 26.61 \\ 
CG (softmax embedding only) & 27.16 \\
CG (both)& \textbf{27.75} \\ \bottomrule
\end{tabular}
\end{center}
\caption{Experiments to determine the effectiveness of composition based embeddings and gated embeddings. We used en-de language pair from the TED multi-target dataset. Std. is our baseline with standard word embeddings, model C is the composition only model and CG  combines the character-aware (composed) embedding and standard embedding via a gating function. 
}
\label{tab:prelim}
\end{table}
\subsection{Large Vocabulary Approximation}\label{sec:approx}
In Equation~\ref{eq:outputdist} of the general NMT framework, the softmax operation generates a distribution over the output vocabulary.
Our character-aware model requires a much larger computation graph as we apply convolutions (and highway layers) over the spellings (character embeddings) of entire target vocabulary, placing a limitation on the target vocabulary size for our model.
Which is problematic for word-level modeling (without BPE).

To make our character-aware model accommodate large target vocabulary sizes, we incorporate an approximation mechanism based on~\cite{chousing}. 
Instead of computing the softmax over the entire vocabulary, 
we uniformly sample $20$k vocabulary types and the vocabulary types that are present in the training batch.

During decoding, we compute the forward pass $\vec{W_o}\vec{s}_j$ in Equation~\ref{eq:outputdist} in several splits of the target vocabulary.
As no backward pass is required we clear the memory (i.e.\ delete the computation graph) after each split is computed.
\section{Experiments} \label{sec:exp}
\begin{table*}[t]
\small
\centering
\begin{tabular}{@{}cccccccccc@{}}
\toprule  
\multicolumn{1}{l}{\textbf{Language}} & \multicolumn{3}{c|}{\textbf{BPE Sweep}} & \multicolumn{3}{c|}{\textbf{@ $30$k BPE}}& \multicolumn{3}{c}{\textbf{@ Word-level}}\\ 
\multicolumn{1}{c}{} & \textbf{Std}(Best BPE) & \textbf{CG}(Best BPE) & \multicolumn{1}{c|}{\textbf{$\Delta$}} & \textbf{Std} & \textbf{CG} & \multicolumn{1}{c|}{\textbf{$\Delta$}}& \textbf{Std} & \textbf{CG} & \textbf{$\Delta$}\\ \midrule
cs & 20.57 (7.5k) & 21.41 (7.5k) & \multicolumn{1}{l|}{+0.84}& 18.73 & 21.28 & \multicolumn{1}{l|}{+2.55}& 18.44  & 21.49 &  +3.05 \\ 
uk & 15.79 (7.5k) & 16.60 (30k) & \multicolumn{1}{l|}{+0.81}& 14.27& 16.60&\multicolumn{1}{l|}{ +2.33}& 12.94  & 15.30 & +2.36\\ 
pl & 16.76 (15k) & 18.00 (30k) &  \multicolumn{1}{l|}{+1.24}& 15.98& 18.00 & \multicolumn{1}{l|}{+2.02}& 15.49  & 17.20& +1.71\\
tr & 15.11 (7.5k) & 15.83 (30k) & \multicolumn{1}{l|}{+0.72}& 13.82& 15.83& \multicolumn{1}{l|}{+2.01}& 12.58 & 14.75& +2.17\\
hu & 16.61 (3.2k) & 17.23 (15k) & \multicolumn{1}{l|}{+0.62}& 15.45& 17.21& \multicolumn{1}{l|}{+1.76}& 14.18& 16.52 & +2.34\\
he & 23.36 (3.2k) & 23.86 (30k) & \multicolumn{1}{l|}{+0.50}& 22.47 & 23.86 & \multicolumn{1}{l|}{+1.39} & 21.26 & 23.01 & +1.75\\ 
pt & 37.85 (15k) & 38.35 (30k) & \multicolumn{1}{l|}{+0.50} & 37.05 & 38.35 & \multicolumn{1}{l|}{+1.30} & 37.13 & 38.36& +1.23\\
ar & 16.22 (7.5k) & 16.28 (30k) & \multicolumn{1}{l|}{+0.06}& 15.05& 16.28& \multicolumn{1}{l|}{+1.23}& 14.45& 16.05&+1.60 \\  
de & 27.37 (7.5k) & 28.12 (30k) & \multicolumn{1}{l|}{+0.75}& 26.94& 28.12& \multicolumn{1}{l|}{+1.21} & 26.84 & 27.75 & +0.91\\ 
ro & 24.02 (3.2k) & 24.20 (15k) & \multicolumn{1}{l|}{+0.18}& 22.88& 24.00& \multicolumn{1}{l|}{+1.12} &22.39 & 23.27& +0.88\\ 
bg & 31.63 (7.5k) & 32.20 (15k) & \multicolumn{1}{l|}{+0.57}& 30.92& 31.90& \multicolumn{1}{l|}{+0.98}& 30.18& 31.43& +1.25\\
fr & 35.97 (1.6k) & 36.17 (7.5k) & \multicolumn{1}{l|}{+0.20}& 35.31& 35.92& \multicolumn{1}{l|}{+0.61}& 35.28 & 36.01  & +0.73\\ 
fa & 12.94 (30k) & 13.52 (30k) & \multicolumn{1}{l|}{+0.58}& 12.94 & 13.52& \multicolumn{1}{l|}{+0.58}& 12.85 & 12.79& -0.06\\
ru & 19.28 (30k) & 19.61 (30k) & \multicolumn{1}{l|}{+0.33}& 19.28 & 19.61& \multicolumn{1}{l|}{+0.33}& 17.60& 19.04& +1.44\\ 
\bottomrule
\end{tabular}
\caption{Best BLEU scores swept over $6$ different BPE merge setting ($1.6$k, $3.2$k, $7.5$k, $15$k, $30$k, $60$k), and at a standard setting of $30$k. We notice a consistent improvement across languages and settings of the merge operation parameter.
}
\label{tab:avg_bpe}
\end{table*}
We evaluate our character aware model on $14$ different languages in a low-resource setting. Additionally, we sweep over several BPE merge hyperparameter settings from character-level to fully word-level for both our model and the baseline and find consistent gains in the character-aware model over the baseline. These gains are stable across all BPE merge hyperparameters all the way up to word-level where they are the highest.
\subsection{Datasets}
We use a collection of TED talk transcripts \cite{duh18multitarget,mauro2012wit3}.
This dataset has languages with a variety of morphological typologies, which allows us to observe how the success of our character-aware decoder relates to morphological complexity.
We keep the source language fixed as English and translate into $14$ different languages, since our focus is on the decoder. 
The training sets for each vary from 74k sentences pairs for Ukrainian to around 174k sentences pairs for Russian (provided in Appendix A), but the validation and test sets are ``multi-way parallel'', meaning the English sentences (the source side in our experiments) are the same across all $14$ languages, and are about $2$k sentences each.
We filter out training pairs where the source sentence was longer that $50$ tokens (before applying BPE). For word-level results, we used a vocabulary size of $100$k (keeping the most frequent types) and replaced rare words by an \texttt{<UNK>} token. 
\subsection{NMT Setup}
We work with OpenNMT-py~\cite{klein2017OpenNMT}, and modify the target-side embedding layer and softmax layer to use our proposed character-aware composition function.
A $2$ layer encoder and decoder, with $1000$ recurrent units were used in all experiments
The embeddings sizes were made to match the RNN recurrent size.
We set the character embedding size to $50$ and use four CNNs with kernel widths $3, 4, 5$ and $6$. The four CNN outputs are concatenated into a compositional embeddings and gated with a standard word embedding.
The same composition function (with shared parameters) was used for the target embedding layer and the softmax layer.

We optimize the NMT objective (Equation~\ref{eq:objective}) using SGD.\footnote{SGD outperformed both Adam and Adadelta. Others have found similar trends, see \newcite{bahar2017empirical} and \newcite{maruf2017document}.}
An initial learning rate of 1.0 was used for the first $8$ epochs and then decayed with a decay rate of $0.5$ until the learning rate reached a minimum threshold of $0.001$.
We use a batch size of 80 for our main experiments. 
At the end of each epoch we checkpoint and evaluate our model on a validation datset and used validation accuracy as our model selection criteria for test time.
During decoding, a beam size of $5$ was chosen for all the experiments.
\subsection{Results}

\begin{table}[t]
\small
\centering
\begin{tabular}{@{}lcccc@{}}
\toprule
\multicolumn{1}{@{}l@{}}{\multirow{2}{*}{\textbf{Lang}}}  & \multicolumn{1}{c}{\multirow{2}{*}{\textbf{\begin{tabular}[c]{@{}c@{}}Char-\\ Shallow\end{tabular}}}} & \multicolumn{1}{c}{\multirow{2}{*}{\textbf{\begin{tabular}[c]{@{}c@{}}Char-\\ Deep\end{tabular}}}} &  
\multicolumn{1}{c}{\multirow{2}{*}{\begin{tabular}[c]{@{}c@{}}\textbf{CG}\\ \textbf{($30$k BPE)}\end{tabular}}} & \multicolumn{1}{@{}c@{}}{\multirow{2}{*}{\textbf{$\Delta$}}}\\
\multicolumn{1}{c}{} & \multicolumn{1}{c}{} & \multicolumn{1}{c}{} & \multicolumn{1}{c}{} & \multicolumn{1}{c}{} \\ \midrule
uk & \phantom{1}4.77& 13.34 & 16.60 & +3.26\\
cs & 11.16& 18.45 & 21.28 & +2.83\\
de &23.89&25.93& 28.12 & +2.19 \\
bg & 26.40&29.81& 31.90 & +2.09\\
tr & \phantom{1}5.29 & 13.94 & 15.83 & +1.89\\
pl & 10.65 & 16.31& 18.00 & +1.69\\
ru &14.63&18.01& 19.61 & +1.60\\
ro &21.58&22.45&24.00 & +1.55\\
pt &35.00& 37.06& 38.35 & +1.29\\
hu & \phantom{1}2.51& 16.02 & 17.21 & +1.19\\
fr & 32.71& 34.76 &35.92 & +1.16\\
fa & \phantom{1}7.44&12.73 & 13.52 & +0.79\\
ar & \phantom{1}3.58&15.89& 16.28 & +0.39\\
he & 22.28&23.87& 23.86 & -0.01\\
\bottomrule
\end{tabular}
\caption{BLEU scores (lowercased) comparing character-level models against CG when used on $30$k BPE sequences. We show that without sweeping BPE, CG generally outperforms purely character-level methods, even when the purely character-level networks are deepened as was shown to help in~\newcite{cherry2018revisiting}.
}
\label{tab:chars}
\end{table}
We provide case insensitive BLEU scores for our main experiments, comparing our character-aware model (CG) 
against a baseline model that uses only standard word (and subword) embeddings. 
We divide the results of our model's performance into three parts: 
\begin{enumerate*}[label=(\roman*)]
\item over a sweep of BPE merge operations, including a commonly used setting of $30$k merge operations 
\item with word-level source and target sequences 
and finally, \item against a purely character-level model. 
\end{enumerate*}
\subsubsection{BPE Results}
Part $1$ of Table~\ref{tab:avg_bpe} compares the best BLEU score obtained by the baseline model, after performing a BPE sweep from $1.6$k to $60$k, to the best BLEU obtained by CG after sweeping over the same BPE range. While our study focuses on the target side, BPE (with the same number of merge operations) was applied to both source and target for our experiments.
We find that after this sweep, CG outperforms the baseline in all 14 languages.
The exhaustive table of results for these experiments is presented in Appendix \ref{detailed_results}. 

\subsubsection*{No Typical BPE Setting}
Additionally, we see that the BPE setting that achieves best BLEU in the baseline model varies considerably from $1.6$k  to $30$k depending on the target language, indicating that \emph{there is no ``typical'' BPE for low-resource settings}. In the CG model, however, performance was usually best at $30$k.
Part $2$ of Table~\ref{tab:avg_bpe} compares the baseline and CG at BPE of $30$k where CG performs optimally. 

We find that our CG model consistently outperforms the baseline for almost all BPE merge hyperparameters across all 14 languages. 
Figure~\ref{fig:deltableu} shows the gains observed by the CG model as we sweep over BPE merge operations. While the baseline model does slightly better than CG at small BPE settings for a few languages (all points below the $0$ value), a majority of the points show positive gains.
\subsubsection{Word-Level Results}
In Part $3$ of Table \ref{tab:avg_bpe} we show results with our approximation for word level. While our best results are generally with BPE, we note that we get the biggest relative gains using our method at the word level, which we expect is due to always having the whole word to learn character patterns over. 
For the CG model, in $60$k BPE and word-level settings we used the large vocabulary approximation discussed in Section~\ref{sec:approx}.
\subsubsection{Character-Level Results}
Finally, in Table \ref{tab:chars}, we compare two character-level models against our CG model at $30$k BPE. The shallow character-level model used $2$ encoder and decoder layers with $1000$ recurrent units, while the deep model used $6$ encoder and decoder layers with $512$ recurrent units .\footnote{Increasing the recurrent size for deep models resulted in significant drop in BLEU scores. We set the dropout rate to $0.1$.} Furthermore, the improved results from the deep model were only attainable using the Fairseq toolkit with Noam optimization and $100$ warmup steps~\cite{gehring2017convs2s}. As Table~\ref{tab:chars} shows, our CG model with $30$k BPE compares favorably to even deep character-level models for this low-resource setting.

\definecolor{col1}{RGB}{76, 114, 176}
\definecolor{col2}{RGB}{221, 132, 82}
\definecolor{col3}{RGB}{85, 167, 104}
\definecolor{col4}{RGB}{196, 78, 82}
\definecolor{col5}{RGB}{129, 114, 178}
\definecolor{col6}{RGB}{147, 120, 96}
\definecolor{col7}{RGB}{218, 139, 195}
\definecolor{col8}{RGB}{140, 140, 140}
\definecolor{col9}{RGB}{204, 185, 116}
\definecolor{col10}{RGB}{100, 181, 205}
\definecolor{col11}{RGB}{198, 219, 239}
\definecolor{col12}{RGB}{158, 202, 225}
\definecolor{col13}{RGB}{107, 174, 214}
\definecolor{col14}{RGB}{49, 130, 189}
\begin{figure}
\centering
\small
\resizebox {\columnwidth} {!} {
\begin{tikzpicture}
    \begin{axis}[
    	ylabel style={at={(axis description cs:+0.15,.5)},anchor=south,align=center,font=\small},
        xlabel style={font=\small},
    	legend pos=outer north east,
        xmin=-0.35,
        xmax=6.35, 
        xlabel=BPE Merge Operations,
        ylabel=$\Delta$ BLEU\\(CG - Std.),
        xticklabels={0, 1.62k, 3.2k, 7.5k, 15k, 30k, 60k$^*$, W$^*$}]

    \addplot[smooth,color=col1, opacity=0.4,thin ]
        plot coordinates {
            (0, 0.43) (1, 0.52) (2, 0.84) (3, 1.54) (4, 2.55) (5, 3.37) (6, 3.05)
        };
    \addplot[clip mode=individual, only marks, color=col1, mark=text, text mark=cs, text mark as node, text mark style={font=\tiny}]
        plot coordinates {
           (0, 0.43) (1, 0.52) (2, 0.84) (3, 1.54) (4, 2.55) (5, 3.37) (6, 3.05)
        };
   \addplot[smooth,color=col2, opacity=0.4]
        plot coordinates {
            (0, 0.45) (1, 0.65) (2, -0.31) (3, 0.92) (4, 2.33) (5, 3.04) (6, 2.36)
        };
   \addplot[clip mode=individual, only marks, color=col2, mark=text, text mark=uk, text mark as node, text mark style={font=\tiny}]
        plot coordinates {
           (0, 0.45) (1, 0.65) (2, -0.31) (3, 0.92) (4, 2.33) (5, 3.04) (6, 2.36)
        };
    \addplot[smooth,color=col3, opacity=0.4,thin ]
        plot coordinates {
            (0, 0.81) (1, 0.28) (2, 1.26) (3, 0.62) (4, 1.76) (5, 2.24) (6, 2.34)
        };
    \addplot[clip mode=individual, only marks, color=col3, mark=text, text mark=hu, text mark as node, text mark style={font=\tiny}]
        plot coordinates {
           (0, 0.81) (1, 0.28) (2, 1.26) (3, 0.62) (4, 1.76) (5, 2.24) (6, 2.34)
        };
    \addplot[smooth,color=col4, opacity=0.4,thin]
        plot coordinates {
             (0, 0.74) (1, 0.72) (2, 0.50) (3, 0.87) (4, 2.02) (5, 1.85) (6, 1.71)
        };
    \addplot[clip mode=individual, only marks, color=col4, mark=text, text mark=pl, text mark as node, text mark style={font=\tiny}]
        plot coordinates {
          (0, 0.74) (1, 0.72) (2, 0.50) (3, 0.87) (4, 2.02) (5, 1.85) (6, 1.71)
        };
    \addplot[smooth,color=col5, opacity=0.4 ,thin]
        plot coordinates {
             (0, 0.45) (1, 0.02) (2, 0.33) (3, 0.57) (4, 1.39) (5, 0.94) (6, 1.75)
        };
    \addplot[clip mode=individual, only marks, color=col5, mark=text, text mark=he, text mark as node, text mark style={font=\tiny}]
        plot coordinates {
          (0, 0.45) (1, 0.02) (2, 0.33) (3, 0.57) (4, 1.39) (5, 0.94) (6, 1.75)
        };
    \addplot[smooth,color=col6, opacity=0.4 ,thin]
        plot coordinates {
             (0, -0.50) (1, 0.67) (2, 0.40) (3, 0.79) (4, 2.01) (5, 1.36) (6, 2.17)
        };
    \addplot[clip mode=individual, only marks, color=col6, mark=text, text mark=tr, text mark as node, text mark style={font=\tiny}]
        plot coordinates {
          (0, -0.50) (1, 0.67) (2, 0.40) (3, 0.79) (4, 2.01) (5, 1.36) (6, 2.17)
        };
    \addplot[smooth,color=col7, opacity=0.4,thin ]
        plot coordinates {
             (0, 0.30) (1, -0.12) (2, -0.05) (3, 0.29) (4, 1.23) (5, 0.67) (6, 1.60)
        };
    \addplot[clip mode=individual, only marks, color=col7, mark=text, text mark=ar, text mark as node, text mark style={font=\tiny}]
        plot coordinates {
             (0, 0.30) (1, -0.12) (2, -0.05) (3, 0.29) (4, 1.23) (5, 0.67) (6, 1.60)
        };
    \addplot[smooth,color=col8, opacity=0.4 ,thin]
        plot coordinates {
            (0, 0.47) (1, 0.45) (2, 0.16) (3, 0.43) (4, 1.30) (5, 1.00) (6, 1.23)
        };
    \addplot[clip mode=individual, only marks, color=col8, mark=text, text mark=pt, text mark as node, text mark style={font=\tiny}]
        plot coordinates {
             (0, 0.47) (1, 0.45) (2, 0.16) (3, 0.43) (4, 1.30) (5, 1.00) (6, 1.23)
        };
 \addplot[smooth,color=col9, opacity=0.4 ,thin]
        plot coordinates {
            (0, 0.07) (1, -0.60) (2, -0.11) (3, 0.42) (4, 1.12) (5, 0.65) (6, 0.88)
        };
  \addplot[clip mode=individual, only marks, color=col9, mark=text, text mark=ro, text mark as node, text mark style={font=\tiny}]
        plot coordinates {
             (0, 0.07) (1, -0.60) (2, -0.11) (3, 0.42) (4, 1.12) (5, 0.65) (6, 0.88)
        };
 \addplot[smooth,color=col10, opacity=0.4 ,thin]
        plot coordinates {
            (0, 0.26) (1, 0.30) (2, 0.18) (3, 1.11) (4, 0.98) (5, 1.14) (6, 1.25)
        };
  \addplot[clip mode=individual, only marks, color=col10, mark=text, text mark=bg, text mark as node, text mark style={font=\tiny}]
        plot coordinates {
             (0, 0.26) (1, 0.30) (2, 0.18) (3, 1.11) (4, 0.98) (5, 1.14) (6, 1.25)
        };
     \addplot[smooth,color=col1, opacity=0.4,thin]
        plot coordinates {
          (0, 0.51) (1, 0.55) (2, 0.35) (3, 0.50) (4, 0.33) (5, 0.95) (6, 1.44)
        };
    \addplot[clip mode=individual, only marks, color=col1, mark=text, text mark=ru, text mark as node, text mark style={font=\tiny}]
        plot coordinates {
             (0, 0.51) (1, 0.55) (2, 0.35) (3, 0.50) (4, 0.33) (5, 0.95) (6, 1.44)
        };
    \addplot[smooth,color=col2, opacity=0.4 ,thin]
        plot coordinates {
          (0, -0.04) (1, 0.21) (2, 0.09) (3, 0.66) (4, 1.18) (5, 0.16) (6, 0.91)
        };
    \addplot[clip mode=individual, only marks, color=col2, mark=text, text mark=de, text mark as node, text mark style={font=\tiny}]
        plot coordinates {
             (0, -0.04) (1, 0.21) (2, 0.09) (3, 0.66) (4, 1.18) (5, 0.16) (6, 0.91)
        };

    \addplot[smooth,color=col3, opacity=0.4 ,thin]
        plot coordinates {
         (0, -0.52) (1, 0.27) (2, 0.52) (3, 0.42) (4, 0.58) (5, 0.11) (6, -0.06)
        };
    \addplot[clip mode=individual, only marks, color=col3, mark=text, text mark=fa, text mark as node, text mark style={font=\tiny}]
        plot coordinates {
             (0, -0.52) (1, 0.27) (2, 0.52) (3, 0.42) (4, 0.58) (5, 0.11) (6, -0.06)
        };
    \addplot[smooth,color=col4, opacity=0.4,thin ]
        plot coordinates {
        (0, -0.08) (1, -0.07) (2, 0.35) (3, 0.20) (4, 0.61) (5, 0.75) (6, 0.73)
        };
    \addplot[clip mode=individual, only marks, color=col4, mark=text, text mark=fr, text mark as node, text mark style={font=\tiny}]
        plot coordinates {
             (0, -0.08) (1, -0.07) (2, 0.35) (3, 0.20) (4, 0.61) (5, 0.75) (6, 0.73)
        };
    \addplot[smooth,color=Black, opacity=1.0, ultra thick,mark=*,mark size=1pt]
        plot coordinates {
        (0, 0.24)  (1, 0.28)  (2, 0.32)  (3, 0.67)  (4, 1.39)  (5, 1.3)   (6, 1.53)
        };
    \addplot[mark=none, Black, samples=2, domain=-1:7, thin] {0.0};
    \end{axis}
    \end{tikzpicture}
    }
    \caption{Plot of the difference between the BLEU scores from CG model and baseline model at various BPE settings for each of the $14$ languages (shown in color, with language identifier). The bold black line shows the average difference across the languages for each BPE setting.}
    \label{fig:deltableu}
\end{figure}
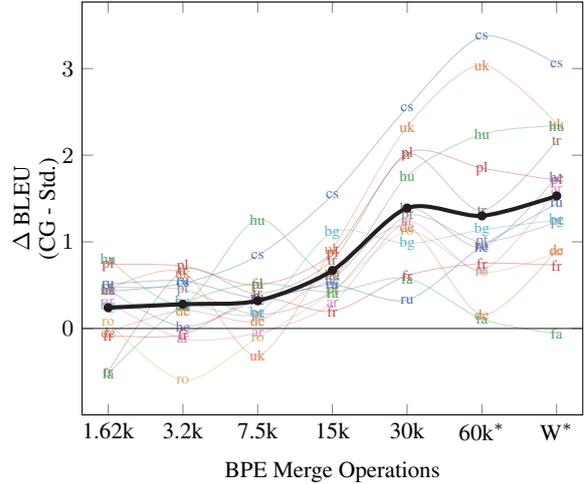
\section{Analysis}\label{sec:analysis}
\begin{table}[]
\resizebox{\columnwidth}{!}{
\small
\centering
\begin{tabular}{@{}p{1.75cm}lccccc@{}}
\toprule
\multirow{2}{*}{\textbf{Features}} & \multicolumn{3}{c}{\textbf{\begin{tabular}[c]{@{}c@{}}Corpus-\\ dependent\end{tabular}}} & \multicolumn{2}{c}{\textbf{\begin{tabular}[c]{@{}c@{}}Corpus-\\ independent\end{tabular}}} \\
 & \textbf{TT} & \textbf{A} & \textbf{H} & \textbf{UT} & \textbf{UTC} \\ \midrule
\textbf{Correlation} & 0.04 & 0.59 & 0.67 & 0.80 & 0.49 \\ \bottomrule
\end{tabular}}
\caption{The Pearsons correlation between the features and the relative gain in BLEU obtained by the CG model. See Section~\ref{sec:analysis} for details regarding features.} \label{tab:correlation}
\end{table}
We are interested in understanding whether our character-aware model is exploiting morphological patterns in the target language.
We investigate this by inspecting the relationship between a set of hand-picked features and improvements obtained by our model over the baseline at word-level inputs.
These features fall into two categories, \emph{corpus-dependent} and \emph{corpus-independent}.
We following~\newcite{bentz2016comparison}, and extract features known to correlate with human judgments of morphological complexity.
The following corpus-dependent features were used:
\begin{enumerate}[label=(\roman*)]
  \item Type-Token Ratio (TT): the ratio of the number of word types to the total number of word tokens in the target side. 
 We note that a large corpus tends to have a smaller type-token ratio compared to small corpus.
  \item Word-Alignment Score (A): computed as $A = \frac{\mid\text{many-to-one} \mid - \mid \text{one-to-many} \mid} { \mid \text{all-alignments} \mid}$. One-to-one, one-to-many and many-to-one alignment types are  illustrated in Figure~\ref{fig:alignments}.\footnote{We use FastAlign~\cite{dyer2013simple} for word alignments with the grow-diag-final-and heuristic from~\cite{och2003systematic} for symmetrization.}
We intuit that a morphologically 
poor source language (like English) paired with a richer target language should exhibit more many-to-one alignments---a single word in the target will contain more information (via morphological phenomena) that can only be translated using multiple words in the source.
  \item Word-Level Entropy (H): computed as $H = \sum_{v \in \mathcal{V}} p(v) \log p(v)$ where $v$ is a word type. This metric reflects the average information content of the words in a corpus. Languages with more dependence on having a large number of word types rather than word order or phrase structure will score higher. 
\end{enumerate}
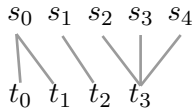
\begin{figure}[h]
\centering
\tikzset{every picture/.style={line width=1.00pt}} 
\begin{tikzpicture}[x=0.75pt,y=0.75pt,yscale=-1,xscale=1]
\draw [color={rgb, 255:red, 155; green, 155; blue, 155 }  ,draw opacity=1 ][line width=1.00]    (90,21.5) -- (70,47) ;
\draw [color={rgb, 255:red, 155; green, 155; blue, 155 }  ,draw opacity=1 ][line width=1.00]    (70,21.5) -- (70,47) ;
\draw [color={rgb, 255:red, 155; green, 155; blue, 155 }  ,draw opacity=1 ][line width=1.00]    (51,22) -- (70,47) ;
\draw [color={rgb, 255:red, 155; green, 155; blue, 155 }  ,draw opacity=1 ]   (46.5,45.5) -- (31,22) ;
\draw [color={rgb, 255:red, 155; green, 155; blue, 155 }  ,draw opacity=1 ]   (26.5,46) -- (8,21) ;
\draw [color={rgb, 255:red, 155; green, 155; blue, 155 }  ,draw opacity=1 ]   (10,45.5) -- (8,21) ;
\draw (30,12) node   {$s_{1}$};
\draw (10,12) node   {$s_{0}$};
\draw (50,12) node   {$s_{2}$};
\draw (70,12) node   {$s_{3}$};
\draw (90,12) node   {$s_{4}$};
\draw (30,52) node   {$t_{1}$};
\draw (10,52) node   {$t_{0}$};
\draw (50,52) node   {$t_{2}$};
\draw (70,52) node   {$t_{3}$};
\end{tikzpicture}
\caption{Example of one-to-many ($s_0$ to $t_0,t_1$), one-to-one ($s_1$ to $t_2$) and many-to-one ($s_2,s_3,s_4$ to $t_3$) alignments. For this example $A = (3-2)/6$.}
\label{fig:alignments}
\end{figure}

For the corpus-independent features we used a morphological annotation corpus called UniMorph~\cite{sylak2015universal}.
The UniMorph corpus contains a large list of inflected words (in several languages) along with the word's lemma and a set of morphological tags.
For example, the French UniMorph corpus contains the word \texttt{marchai} (walked), which is associated with its lemma, \texttt{marcher} and a set of morphological tags \textbf{$\{$\texttt{V,IND,PST,1,SG,PFV}$\}$}. 
There are $19$ such tags in the French UniMorph corpus.
A morphologically richer language like Hungarian, for example, has $36$ distinct tags.
We used the number of distinct tags (UT) and the number of different tag combinations (UTC) that appear in the UniMorph corpus for each language.
Note that we do not filter out words (and its associated tags) from the UniMorph corpus that are absent in our parallel data. 
This ensures that the UT and UTC features are completely corpus independent.

The Pearson's correlation between these hand-picked features and relative gain observed by our model is shown in Table~\ref{tab:correlation}.
For this analysis we used the relative gain obtained from the word-level experiments.
Concretely, the relative gain for Czech was computed as $\frac{21.49 - 18.44}{18.44}$
We see a strong correlation between the corpus-independent feature (UT) and our model's gain. Alignment score and Word Entropy are also moderately correlated. Surprisingly, we see no correlation to type-token ratio.

As the correlation analysis only examines the relation between BLEU gains and an \emph{individual} feature, we further analyzed how the features \emph{jointly} relate to BLEU gains.
We fitted a linear regression model, setting the relative gains as the predicted variable $y$ and the feature values as the input variables $\vec{x}$, with the goal of studying the linear regression weights $\vec{\phi}$.\footnote{The input features were min-max normalized for the regression analysis.}
We used feature-augmented domain adaptation 
where we consider each language as a domain~\cite{daume2007frustratingly}, allowing the model to find a set of ``general'' weights as well language-specific weights that best fit the data (Equation~\ref{eq:lrobj}). 
The general feature weights can be interpreted as being indicative of the overall trends in the dataset across all the languages, while the language-specific weights indicate
language deviation from the overall trend. 
\begin{align}
  \vec{\mathcal{L}}(\vec{\phi}) &= \sum_{i \in \mathcal{I}} \mid y_i - \tilde{y}_i \mid^2  - \lambda \mid \vec{\phi} \mid^2\\ \label{eq:lrobj}
  \tilde{y}_i&= \vec{\phi}_{\text{ALL}}^T\vec{x}_i + \vec{\phi}_i^T\vec{x}_i 
\end{align}
Where, $y$ is the true relative gain in BLEU, $\tilde{y}$ is the predicted gain, $\vec{x}$ is a vector of input feature values, $\vec{\phi}_{\text{ALL}}$ and $\vec{\phi}_i$ are the general and language-specific weights, and $i$ indexes into the set of languages in our analysis.
We set $\lambda$ to $0.05$.
\begin{figure}
\centering
  \includegraphics[width=0.9\columnwidth]{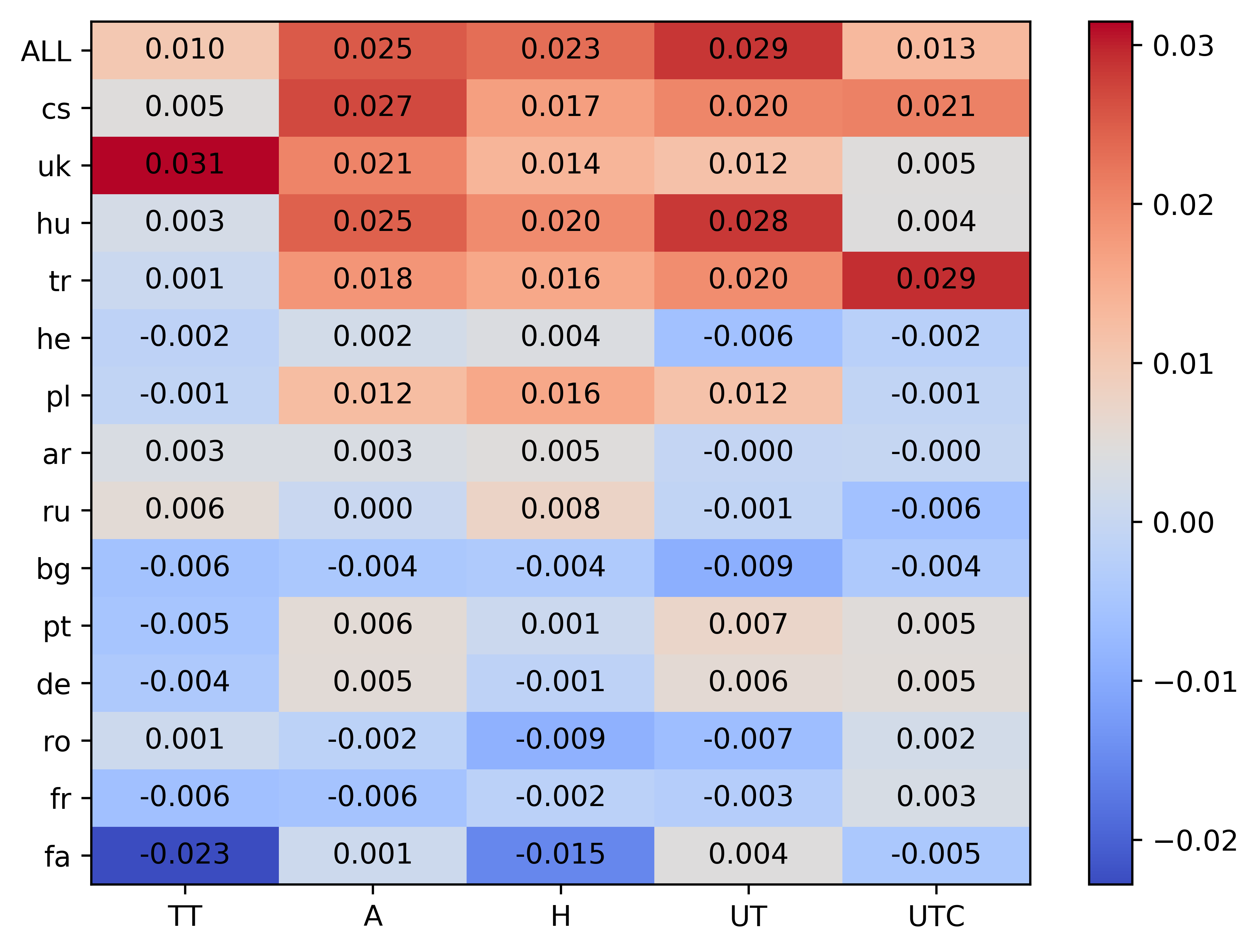}
\caption{Feature weights of the feature-augmented language adapted linear regression model. The first row represents the ``general'' set of weights used for all of the languages.
Each row below are the language-adapted weights that only ``fire'' for that specific language.}
\label{fig:lrweights}
\end{figure}

The matrix of learned weights $\vec{\phi}$ is visualized in Figure~\ref{fig:lrweights}.
The first row of weights correspond to the ``general'' weights that are used for all the languages, followed by language-specific weights sorted by relative gain.

While the general weights align with the correlation results (Table~\ref{tab:correlation}),
this analysis also shows that the UTC weight for Czech and Turkish are much larger than any of the other languages' and indeed we can verify that these languages have $194$ and $300$ different tag combinations while the average tag combinations is $\approx110$.

From the corpus-dependent features, word alignment score strongly predicts the gain in BLEU scores. 
For Czech, Ukrainian, Turkish, Hungarian, and Polish we see additional weight placed on this feature. 
A similar trend can be seen for the word-entropy feature.
While type-token ratio does not exhibit a strong overall trend, 
we see that Ukrainian and Farsi are outliers. 


Our correlation and regression analysis strongly suggest that CG character-aware modeling helps the most when the target language has inherent morphological complexity and that it does indeed have the ability to handle morphological patterns present in the target languages.
\subsection{Qualitative Examples}

\begin{table}
\small
\begin{center}
\begin{tabular}{ |l|l| } 
 \hline
Src & here he is : leonardo \textbf{da vinci} .  \\  
Ref & h*A hw -- lywnArdw \textbf{dA fyn\$y} .\\  
Std &  hnA hw : lywnArdw \textbf{dA dA} .\\  
CG &  hnA hw : lywnArdw \textbf{dA fy+n\$y} .\\  \hline  
Src &  i 'm the \textbf{mexican} in the family . \\  
Ref & AnA \textbf{Almksyky} fy AlEA\}lp .\\  
Std &  AnA \textbf{mksy+Any} fy AlEA\}lp . \\  
CG &  AnA \textbf{Almksy+ky} fy AlEA\}lp . \\  \hline  
Src &  there was going to be a national \textbf{referendum} . \\  
Ref & wtm AlAEdAd lAHrA' \textbf{AstftA'} \$Eby .\\  
Std & sykwn hnAk \textbf{f+tA'} wTny .\\  
CG & sykwn hnAk \textbf{Ast+f+tA'} wTny . \\
 \hline
Src & there are ordinary \textbf{heroes} . \\  
Ref & fhnAk \textbf{AbTAl} TbyEywn .\\  
Std & hnAk \textbf{ASdqA'} EAdy .\\  
CG &  hnAk \textbf{AbTAl} EAdyyn .\\
 \hline
\end{tabular}
\end{center}
\caption{Examples from En-Ar, transliterated with the Buckwalter schema. We show the version of our model and the English using `+' to denote where BPE splits words up, while BPE has not been applied to the target reference.
\Adi{room for one or two more example?}}
\label{tab:examples}
\end{table}

We additionally look at specific examples of where our model is outperforming the baseline in the case of $30$k BPE in En-Ar. We see a few trends, which we show examples of in Table \ref{tab:examples}. The first trend, corresponding to the first example, is that it gets names better. This might be because Arabic is not written in the Latin alphabet, and the spelling-aware model may be able to transliterate better. 

Another trend is that CG gets the endings of rare words correct, in particular when the BPE segmentation is \emph{not} according to morpheme boundaries. The second example illustrates this, where the word for ``Mexican'' appears in the training data broken up by BPE with various morphological endings, all of which are spelled beginning with ``ky'' in the second subword. The morpheme boundaries here would be ``Al+mksyk+y.'' Note that CG also gets the definite article ``Al'' correct while the baseline does not.

Finally, we see a pattern where our model does better for words which are rare and appear both with and without the definite article ``Al.'' Our third example in Table \ref{tab:examples} illustrates this with an infrequent word, the word for ``referendum'', which gets broken up into subwords. In particular, the first subword sometimes has an ``Al'' attached in the training data. Our model is able to translate this subword, while the baseline skips the subword altogether, outputting two subwords that alone are not a valid word. Again, the word is not broken up along morpheme boundaries by BPE. Here there would be no way to break this word up into morphological segments---it consists of non-concatenative derivational morphology. This occurs again in the fourth example in the word for ``heroes,'' where the baseline predicts the word for ``friends.'' In this case the word was not split up by BPE, but similarly it is rare but occurs with the definite article attached in the training data as well.

\section{Conclusion}
We extend character-aware word-level modeling to the decoder for translation into morphologically rich languages. 
Our improvements were attained by augmenting the softmax and the target embedding layers with character-awareness.
We also find it critical to add a gating function to balance compositional embeddings with standard embeddings.
We evaluate our method on a low-resource dataset translating from English into 14 languages, and on top of a spectrum of BPE merge operations. Furthermore, for word-level and higher merge hyperparameter settings, we introduced an approximation to the softmax layer. We achieve consistent performance gains across languages and subword granularities, and perform an analysis indicating that the gains for each language correspond to morphological complexity.

For future work, we would like to explore how our methods might be of use in higher-resource settings. 
Furthermore, it would be interesting to see how these methods might interact with multilingual systems and if they might be able to improve what information is shared between related languages.
\Pamela{I added a sentence to the conclusion to help us hit 8 pages and to end on a bit more of a positive note (and not the one the sassy reviewer commented on)}
\section*{Acknowledgements}
This project originated at the Machine Translation Marathon 2018. We thank the organizers and attendees for their support, feedback and helpful discussions during the event.
This work is supported in part by the Office of the Director of National Intelligence, IARPA. The views contained herein are those of the authors and do not necessarily reflect the position of the sponsors.
\bibliography{mtsummit2019}
\bibliographystyle{mtsummit2019}
\clearpage
\appendix
\section{More Detailed Results} \label{detailed_results}

In Table \ref{tab:data_sizes}, we provide the number of training sentences for each language.

In Table \ref{tab:main_results}, we provide the full experiments of our sweep of BPE for both standard and our CG embeddings. In our baseline, we see a divergence in trends across languages while sweeping over BPE merge hyperparameters---Czech (cs), Turkish (tr), and Ukrainian (uk) for example, are highly sensitive to the BPE merge hyperparameter.
On the other hand, for languages like French (fr) and Farsi (fa), the performance is mostly consistent across different BPE merge hyperparameters.

\Pamela{Do we need to submit this separately?}

\begin{table*}[h]
\centering
\begin{tabular}{lr}
\toprule
Language & Number of sentences \\
\midrule
Czech (cs) & 81k \\
Ukrainian (uk) & 74k\\
Hungarian (hu) & 108k\\
Polish (pl) & 149k\\
Hebrew (he) & 181k\\
Turkish (tr) & 137k\\
Arabic (ar) & 168k\\
Portuguese (pt) & 147k\\
Romanian (ro) & 155k\\
Bulgarian (bg) & 159k\\
Russian (ru) & 174k\\
German (de) & 146k\\
Farsi (fa) & 106k\\
French (fr) & 149k\\
\bottomrule
\end{tabular}
\caption{Number of sentences in training data for each language}
\label{tab:data_sizes}
\end{table*}

\begin{table*}[h]
\small
\centering
\begin{tabular}{@{}clccccccccc@{}}
\toprule
\multicolumn{1}{c}{\multirow{2}{*}{\textbf{L}}} & \multicolumn{1}{c}{\multirow{2}{*}{\textbf{M}}} & \multicolumn{1}{c}{\multirow{2}{*}{\textbf{\begin{tabular}[c]{@{}c@{}}Char-\\ Shallow\end{tabular}}}} & \multicolumn{1}{c}{\multirow{2}{*}{\textbf{\begin{tabular}[c]{@{}c@{}}Char-\\ Deep\end{tabular}}}} &  
\multicolumn{6}{c}{\textbf{BPE (Subwords)}} & \multicolumn{1}{c}{\multirow{2}{*}{\textbf{\begin{tabular}[c]{@{}c@{}}Word-\\ Level\end{tabular}}}} \\
\multicolumn{1}{c}{} & \multicolumn{1}{c}{}  & \multicolumn{1}{c}{} & \multicolumn{1}{c}{} & \textbf{1.6k}& \textbf{3.2k} & \textbf{7.5k} & \textbf{15k} & \textbf{30k} & \textbf{60k} & \multicolumn{1}{c}{} \\ \midrule
\multirow{2}{*}{cs} & Std.& 11.16& 18.45 & 20.28 & 20.51 & 20.57 & 19.60 & 18.73 & 17.60 & 18.44 \\ \vspace{0.07cm}
 & CG &\multicolumn{1}{c}{-}&\multicolumn{1}{c}{-}& 20.71&21.04 & 21.41 & 21.14 & 21.28 & 20.97 & \textbf{21.49} \\ 
 \multirow{2}{*}{uk} & Std. &4.77&-& 13.35 & 15.51 & 15.79 & 15.36 & 14.27 & 12.50 & 12.94 \\ \vspace{0.07cm}
 & CG & \multicolumn{1}{c}{-}&\multicolumn{1}{c}{-}& 13.80& 16.16 & 15.48 & 16.28 & \textbf{16.60} & 15.54 & 15.30 \\
 \multirow{2}{*}{hu} & Std. & 2.51& 16.02 & 15.77 & 16.33 & 15.62 & 16.61 & 15.45 & 14.81 & 14.18 \\ \vspace{0.07cm}
 & CG & \multicolumn{1}{c}{-}&\multicolumn{1}{c}{-}& 16.58 & 16.61 & 16.88 & \textbf{17.23} & 17.21 & 17.05 & 16.52 \\ 
\multirow{2}{*}{pl} & Std. & 10.65&16.31& 16.14& 16.40 & 16.34 & 16.76 & 15.98 & 15.47 & 15.49 \\ \vspace{0.07cm}
 & CG & \multicolumn{1}{c}{-}&  \multicolumn{1}{c}{-}& 16.88 & 17.12 & 16.84 & 17.63 & \textbf{18.00} & 17.32 & 17.20 \\ 
\multirow{2}{*}{he} & Std. & 22.28&23.87&23.07 & 23.36 & 23.32 & 22.76 & 22.47 & 21.84 & 21.26 \\ \vspace{0.07cm}
 & CG & \multicolumn{1}{c}{-}& \multicolumn{1}{c}{-}& 23.52& 23.38 & 23.65 & 23.33 & \textbf{23.86} & 22.78 & 23.01 \\ 
\multirow{2}{*}{tr} & Std. &5.29 & 13.94 & 14.92 & 14.58 & 15.11 & 14.75 & 13.82 & 13.69 & 12.58 \\ \vspace{0.07cm}
 & CG & \multicolumn{1}{c}{-}& \multicolumn{1}{c}{-}& 14.42 & 15.25 & 15.51 & 15.54 & \textbf{15.83} & 15.05 & 14.75 \\ 
\multirow{2}{*}{ar} & Std. & 3.58&15.89& 15.66 &15.67& 16.22 & 15.70 & 15.05 & 14.86 & 14.45 \\ \vspace{0.07cm}
 & CG & \multicolumn{1}{c}{-}& \multicolumn{1}{c}{-}& 15.96 & 15.55 & 16.17 & 15.99 & \textbf{16.28} & 15.53 & 16.05 \\ 
\multirow{2}{*}{pt} & Std. &35.00& 37.06&37.47  & 37.53 & 37.61 & 37.85 & 37.05 & 37.11& 37.13 \\ \vspace{0.07cm}
 & CG & \multicolumn{1}{c}{-}& \multicolumn{1}{c}{-}&37.94& 37.98 & 37.77 & 38.28 & \textbf{38.35} & 38.11 & 38.36 \\ 
\multirow{2}{*}{ro} & Std. &21.58&22.45& 23.48&24.02 & 23.72 & 23.78 & 22.88 & 22.73 & 22.39 \\ \vspace{0.07cm}
 & CG & \multicolumn{1}{c}{-}& \multicolumn{1}{c}{-}& 23.55 & 23.42 & 23.61 & \textbf{24.20} & 24.00 &23.38  & 23.27 \\
\multirow{2}{*}{bg} & Std. & 26.40&29.81& 31.17 & 31.41 & 31.63 & 31.09 & 30.92 & 30.44 & 30.18\\ \vspace{0.07cm}
& CG & \multicolumn{1}{c}{-}& \multicolumn{1}{c}{-}& 31.43 & 31.71 & 31.81 & \textbf{32.20} & 31.90 & 31.58 & 31.43 \\
\multirow{2}{*}{ru} & Std. &14.63&-& 18.17 & 18.71 & 19.05 & 18.80 & 19.28 & 18.28 & 17.60\\ \vspace{0.07cm}
 & CG &\multicolumn{1}{c}{-}& \multicolumn{1}{c}{-}& 18.68 & 19.26 & 19.40 & 19.30 & \textbf{19.61} & 19.23 & 19.04 \\ 
\multirow{2}{*}{de} & Std. &23.89&25.93& 26.98 & 27.34 & 27.37 & 27.23 & 26.94 & 27.21 & 26.84 \\ \vspace{0.07cm}
 & CG & \multicolumn{1}{c}{-}& \multicolumn{1}{c}{-}& 26.94 & 27.55 & 27.46 & 27.89 & \textbf{28.12} & 27.37 &  27.75 \\
\multirow{2}{*}{fa} & Std. & 7.44&12.73 & 12.87 & 12.71 & 12.86 & 12.94 & 12.94 & 13.20 & 12.85 \\ \vspace{0.07cm}
 & CG & \multicolumn{1}{c}{-}& \multicolumn{1}{c}{-}& 12.35 & 12.98 & 13.38 & 13.36 & \textbf{13.52} & 13.31 & 12.79 \\ 
\multirow{2}{*}{fr} & Std. & 32.71& 34.76 & 35.97 & 35.75 & 35.82 & 35.90 & 35.31 & 35.33 & 35.28 \\ \vspace{0.07cm}
 & CG & \multicolumn{1}{c}{-}& \multicolumn{1}{c}{-}& 35.89 & 35.68 & \textbf{36.17} & 36.10 & 35.92 & 36.08 & 36.01 \\ \bottomrule
\end{tabular}\caption{BLEU scores (case insensitive) for a standard embedding encoder-decoder baseline (Std), and character-aware model, composed embedding combined with standard embedding (CG) for $14$ languages and various BPE merge hyperparameters. For purely character-level we only train the standard model as CG would not have a sequence of characters to compose. For BPE of $60$k and word-level we use the softmax approximation described. We see that CG obtains the best result in all languages.
} 
\label{tab:main_results}
\end{table*}
\end{document}